\documentclass[10pt, a4paper]{article}

\usepackage{lrec-coling2024} % this is the new style
\usepackage{tabularx}
\usepackage{booktabs}  
\usepackage{multirow}
\usepackage{seqsplit} 
\usepackage{mdframed}
\usepackage{xcolor}
\usepackage{graphicx}
\usepackage[whole]{bxcjkjatype} % overleafのpdflatexで日本語を使う

\usepackage{url}

\title{Prompting for Numerical Sequences: A Case Study on \\Market Comment Generation}
 \name{Masayuki Kawarada, Tatsuya Ishigaki and Hiroya Takamura} 
\address{Artificial Intelligence Research Center, AIST \\
         masayuki.kawarada.m@gmail.com\\
         \{ishigaki.tatsuya, takamura.hiroya\}@aist.go.jp\\}
\abstract{
Large language models (LLMs) have been applied to a wide range of data-to-text generation tasks, including tables, graphs, and time-series numerical data-to-text settings.
While research on generating prompts for structured data such as tables and graphs is gaining momentum, in-depth investigations into prompting for time-series numerical data are lacking.
Therefore, this study explores various input representations, including sequences of tokens and structured formats such as HTML, LaTeX, and Python-style codes.
In our experiments, we focus on the task of Market Comment Generation, which involves taking a numerical sequence of stock prices as input and generating a corresponding market comment.
Contrary to our expectations, the results show that prompts resembling programming languages yield better outcomes, whereas those similar to natural languages and longer formats, such as HTML and LaTeX, are less effective.
Our findings offer insights into creating effective prompts for tasks that generate text from numerical sequences.
\\ \newline \Keywords{Generation, Data-to-text, Large language model}}

\begin{document}

\maketitleabstract

\section{Introduction}

\begin{figure}[t]
\centering
  \includegraphics[width=\linewidth]{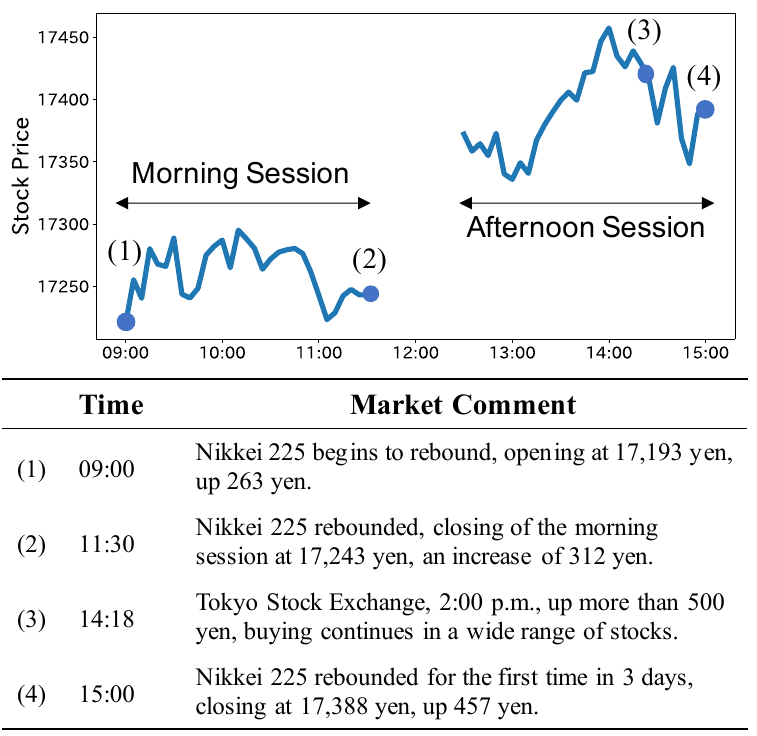}
  \caption{Example of market comment generation. We represent the time-series numerical data using a line graph. The objective of this task is to generate comments on the stock price movement at a specific time.}
  \label{fig:input_format}
\end{figure}

Large language models (LLMs) have demonstrated remarkable performance in various text-to-text natural language generation tasks, such as text summarization~\cite{wang-etal-2023-element, zhang-etal-2023-summit, pham-etal-2023-select}, machine translation~\cite{wang-etal-2023-document-level, karpinska-iyyer-2023-large, vilar2023prompting, agrawal-etal-2023-context}, and dialogue systems~\cite{li-etal-2023-autoconv, jin-etal-2023-instructor}. 
Although LLMs have also been applied to data-to-text generation tasks, their application has been limited to tasks where the input data are structured and their components are represented as words, such as tables~\cite{saha-etal-2023-murmur, zhao-etal-2023-investigating} and structured data~\cite{jiang-etal-2023-structgpt}. 
However, the question of how to effectively use LLMs for text generation from numerical data remains unanswered.
This paper addresses this question through a case study on few-shot market comment generation from stock prices~\cite{murakami-etal-2017-learning, aoki-etal-2018-generating, hamazono-etal-2021-unpredictable}. 
In this task, the stock price is provided as a numerical sequence, and the goal is to generate a market commentary at a specific point in time. 
Figure \ref{fig:input_format} illustrates an example of the market comment generation.

The key challenge in this study is to explore effective ways to represent these numerical sequences as prompts for LLMs, because their characteristics differ significantly from the text used to pretrain the models.
We hypothesize that better performance could be achieved by creating prompts with characteristics similar to those used in the pretraining of LLMs. 
To test this hypothesis, we compare various prompt designs with features that are both similar to and different from the texts used for pretraining.

Specifically, we explore four categories of prompt designs for numerical sequences: 1) directly providing a sequence of numbers; 2) treating the input numerical sequence as a structured table and linearizing it; 3) representing the input sequence as programming code (e.g., Python List, Python Dictionary, HTML, and LaTeX formats), which is a common strategy in code generation research; and 4) using templates to transform the input sequence into natural language.

Contrary to our expectations, our experiments revealed that prompts using expressions closer to programming languages performed better, whereas those closer to natural languages performed worse. 
In addition, longer prompts, such as HTML and LaTeX formats, were found to be less effective. 
These findings suggest that converting numerical data into prompts that closely resemble the format used during pretraining can lead to improved performance.

The main contributions of this study are threefold: 1) we present the performance of zero-shot and few-shot approaches for generating market commentary; 2) we compare various prompting strategies; and 3) we provide insights into the effectiveness of different prompt designs for numerical data.

%To facilitate further research, we provide publicly available pre-processing scripts for the source numerical and textual data\footnote{Obtaining the data itself requires a contract with the content producers. We provide a sample of the preprocessed data in \url{https://github.com/aistairc/market-reporter}}.

\begin{table*}
    \centering
    \footnotesize
    \begin{tabularx}{\textwidth}{lX}
        \toprule
    Direct &  ...9988.05 9982.06 9978.11 9972.66 9967.11, 9961.37 ... \\
    \midrule
    \addlinespace[0.1cm]
    Column &  Time:15:00 14:55 14:50 14:45 14:40 14:35 ... \\
          &  Nikkei225:9988.05 9982.06 9978.11 9972.66 9967.11, 9961.37 ... \\
    Row&  Time Nikkei225\textbackslash{n}15:00 9988.05\textbackslash{n}14:55 9982.06\textbackslash{n}14:50 9978.11\textbackslash{n}14:45 9972.66\textbackslash{n}14:40 9967.11\textbackslash{n}14:35 9961.37 ...\\
    \midrule
    \addlinespace[0.1cm]
    \addlinespace[0.1cm]
    Python List &  Time = [..., "15:00", "14:55", "14:50", "14:45", "14:40", "14:35", ...] \\
                &  Nikkei225 = [..., 9988.05, 9982.06, 9978.11,  9972.66, 9967.11, 9961.37, ...] \\
    \addlinespace[0.1cm]
    Python List~(nested)& Nikkei225 = [..., [15:00, 9988.05], [14:55, 9982.06], [14:50, 9978.11], 14:45, 9972.66], [14:40,9967.11], [14:35,9961.37], ...] \\
    \addlinespace[0.1cm]
    Python Dictionary  &  Nikkei225 = \{..., "15:00":9988.05, "14:55":9982.06, "14:50":9978.11, "14:45":9972.66, "14:40":9967.11, "14:35":9961.37, ...\} \\
    \addlinespace[0.1cm]
    HTML Table & \seqsplit{<table><tr><th>Time</th><th>Nikkei225</th></tr>...(omitted) <tr><td>15:00</td><td>9988.05</td></tr><tr><td>14:55</td><td>9982.06</td></tr><tr><td>14:50</td><td>9978.11</td></tr><tr><td>14:45</td><td>9972.66</td></tr><tr><td>14:40</td><td>9967.11</td></tr><tr><td>14:35</td><td>9961.37</td></tr> ... </table>} \\   
    LaTeX Table &  \textbackslash begin\{table\}[t] \textbackslash begin\{tabular\} \& \textbackslash hline Timestamp \& Nikkei225 \textbackslash \textbackslash \textbackslash hline \textbackslash hline 15:00 \& 9988.05 \textbackslash \textbackslash \textbackslash hline 14:55 \& 9982.06 \textbackslash \textbackslash \textbackslash hline 14:50 \&  9978.11 \textbackslash \textbackslash \textbackslash hline 14:45 \& 9972.66 \textbackslash \textbackslash \textbackslash hline 14:40 \&  9967.11 \textbackslash \textbackslash \textbackslash hline 14:35 \& 9961.37 \textbackslash \textbackslash \textbackslash hline ...
    \textbackslash end\{tabular\}
\textbackslash end\{table\}
    \\
    \midrule
    \addlinespace[0.1cm]
    Text~(English)  & Nikkei225 as of 15:00 is 9982.06 yen.\textbackslash{n}Nikkei225 stock price as of 15:00 is 9988.05 yen.\textbackslash{n}Nikkei225 stock price as of 14:55 is 9982.06 yen.\textbackslash{n}Nikkei225 stock price as of 14:50 is 9978.11 yen.\textbackslash{n}Nikkei225 stock price as of 14:45 is 9972.66 yen.\textbackslash{n}Nikkei225 stock price as of 14:40 is 9967.11 yen.\textbackslash{n}Nikkei225 stock price as of 14:35 is 9961.37 yen....(omitted)
     \\
    \addlinespace[0.1cm]
    Text~(Japanese)  & Nikkei225 as of 15:00 is 9982.06円.\textbackslash{n}15:00時点のNikkei225は9988.05円.\textbackslash{n}14:55時点のNikkei225は9982.06円.\textbackslash{n}14:50時点のNikkei225は9978.11円.\textbackslash{n}14:45時点のNikkei225は9972.66円.\textbackslash{n}14:40時点のNikkei225は9967.11円.\textbackslash{n}14:35時点のNikkei225は9961.37円....(omitted) \\
    \bottomrule
    \end{tabularx}
    \caption{Examples of \texttt{[INPUT FORMAT (short-term)]}.  The actual prompts are written in Japanese except for Text (Japanese).}
    \label{tab:example-input-format-short}
\end{table*}

\begin{table*}
    \centering
    \footnotesize
    \begin{tabularx}{\textwidth}{lX}
        \toprule
    Direct &  9988.05 9982.06 9978.11 9972.66 9967.11, 9961.37 9960.20 \\
    \midrule
    \addlinespace[0.1cm]
    Column &  Date: 7DaysAgo 6DaysAgo 5DaysAgo 4DaysAgo 3DaysAgo 2DaysAgo 1DayAgo \\
          &  Nikkei225: 9988.05 9982.06 9978.11 9972.66 9967.11, 9961.37 9960.20 \\
    Row&  Date Nikkei225\textbackslash{n}7DaysAgo 9988.05\textbackslash{n}6DaysAgo 9982.06\textbackslash{n}5DaysAgo 9978.11\textbackslash{n}4DaysAgo 9972.66\textbackslash{n}3DaysAgo 9967.11\textbackslash{n}2DaysAgo 9961.37 \textbackslash{n}1DayAgo\\
    \midrule
    \addlinespace[0.1cm]
    \addlinespace[0.1cm]
    Python List &  Time = ["7DaysAgo", "6DaysAgo", "5DaysAgo", "4DaysAgo", "3DaysAgo", "2DaysAgo", "1DayAgo"] \\
                &  Nikkei225 = [9988.05, 9982.06, 9978.11,  9972.66, 9967.11, 9961.37, 9960.20] \\
    \addlinespace[0.1cm]
    Python List~(nested)& Nikkei225 = [["7DaysAgo", 9988.05], ["6DaysAgo" 9982.06], ["5DaysAgo", 9978.11], ["4DaysAgo", 9972.66], ["3DaysAgo",9967.11], ["2DaysAgo",9961.37], ...] \\
    \addlinespace[0.1cm]
    Python Dictionary  &  Nikkei225 = \{"7DaysAgo":9988.05, "6DaysAgo":9982.06, "5DaysAgo":9978.11, "4DaysAgo":9972.66, "3DaysAgo":9967.11, "2DaysAgo":9961.37, ... \} \\
    \addlinespace[0.1cm]
    HTML & \seqsplit{<table><tr><th>Date</th><th>Nikkei225</th></tr><tr><td>7DaysAgo</td><td>9988.05</td></tr><tr><td>6DaysAgo</td><td>9982.06</td></tr><tr><td>5DaysAgo</td><td>9978.11</td></tr><tr><td>4DaysAgo</td><td>9972.66</td></tr><tr><td>3DaysAgo</td><td>9967.11</td></tr><tr><td>2DaysAgo</td><td>9961.37</td> ... </tr></table>} \\   
    LaTeX &  \textbackslash begin\{table\}[t] \textbackslash begin\{tabular\} \& \textbackslash hline Timestamp \& Nikkei225 \textbackslash \textbackslash \textbackslash hline \textbackslash hline 7DaysAgo \& 9988.05 \textbackslash \textbackslash \textbackslash hline 6DaysAgo \& 9982.06 \textbackslash \textbackslash \textbackslash hline 5DaysAgo \&  9978.11 \textbackslash \textbackslash \textbackslash hline 4DaysAgo \& 9972.66 \textbackslash \textbackslash \textbackslash hline 3DaysAgo \&  9967.11 \textbackslash \textbackslash \textbackslash hline 2DaysAgo \& 9961.37 \textbackslash \textbackslash \textbackslash hline
    ...
    \textbackslash end\{tabular\}
\textbackslash end\{table\}
    \\
    \midrule
    \addlinespace[0.1cm]
    Text~(English)  & Nikkei225 as of 7 days ago was 9982.06 yen.\textbackslash{n}Nikkei225 closing stock price as of 6 days ago was 9988.05 yen.\textbackslash{n}Nikkei225 closing stock price as of 5 days ago was 9982.06 yen.\textbackslash{n}Nikkei225 closing stock price as of 4 days ago was 9978.11 yen.\textbackslash{n}Nikkei225 closing stock price as of 3 days ago was 9972.66 yen.\textbackslash{n}Nikkei225 closing stock price as of 2 days ago was 9967.11 yen.\textbackslash{n}Nikkei225 closing stock price as of yesterday was 9961.37 yen.
    ...
     \\
    \addlinespace[0.1cm]
    Text~(Japanese)  & 7日前のNikkei225終値は9982.06円.\textbackslash{n}6日前のNikkei225終値は9988.05円.\textbackslash{n}5日前のNikkei225終値は9982.06円.\textbackslash{n}4日前のNikkei225終値は9978.11.\textbackslash{n}3日前のNikkei225終値は9972.66円.\textbackslash{n}2日前のNikkei225終値は9967.11円.\textbackslash{n}2日前のNikkei225終値は9961.37円...
    ...
     \\
    \bottomrule
    \end{tabularx}
    \caption{Examples of \texttt{[INPUT FORMAT (long-term)]}. The actual prompts are written in Japanese except for Text (English).}
    \label{tab:example-input-format-long}
\end{table*}

\begin{table}[t]
  \footnotesize
  \begin{mdframed}
    Output the market comment at the current time in the form of a <comment>market comment</comment>.

   \#\#\#
   
    Input:
    
    \texttt{[INPUT FORMAT (short-term)]}
    
    \texttt{[INPUT FORMAT (long-term)]}
    
    Output: 
    
    \textsl{Nikkei225 closes at large, rebounding yen strength pushes mainstay stocks higher}

    \#\#\#
    
    Input:
    
    \texttt{[INPUT FORMAT (short-term)]}
    
    \texttt{[INPUT FORMAT (long-term)]}
    
    Output:
  \end{mdframed}
  \caption{The template we use for the few-shot setting.}
  \label{tab:prompt-example}
\end{table}

\section{Related Work}

Existing data-to-text generation settings can be categorized based on the input and output types.
The input types include tables~\cite{10.1609/aaai.v33i01.33016908-rotowire,lebret-etal-2016-neural}, graphs~\cite{bai-etal-2022-graph,konstas-etal-2017-neural}, RDF triples~\cite{gardent-etal-2017-webnlg}, and time-series numerical sequence~\cite{murakami-etal-2017-learning,chang-etal-2022-logic,ishigaki-etal-2021-generating, kantharaj-etal-2022-chart}.
From the output types, existing studies generate e.g., bibliographies~\cite{lebret-etal-2016-neural}, summaries of sporting events~\cite{10.1609/aaai.v33i01.33016908-rotowire}, and commentaries~\cite{ishigaki-etal-2021-generating,zhang-etal-2022-moba-commentary,chang-etal-2022-logic}.
Despite the large amount of time-series numerical data in the real world, many existing settings deal with tabular and graphical data owing to public benchmark datasets such as WebNLG, E2E, and RotoWire.
Our study aims to accelerate the research on time-series numerical data.

% numerical sequence-to-textの設定
Various tasks have been proposed as time-series numerical-data-to-text settings, such as commentary generation from user gameplay~\cite{ishigaki-etal-2021-generating}, explaining a drone's movement from sensor data~\cite{chang-etal-2022-logic} and explaining price changes in a market~\cite{murakami-etal-2017-learning,hamazono-etal-2021-unpredictable,aoki-etal-2018-generating}.

% 時系列数値データの問題設定でのモデルの作り方。encoder-decoder vs decoder-only.
% finetuneするモデルがどちらの設定でもある。
% decoder-onlyの場合はzero-shot, few-shotがある。zero-shot, few-shotの研究が活発になりつつあり、調査する価値があることを言う。
Two primary model architectures prevail in data-to-text settings: the encoder-decoder and the decoder-only.
A conventional approach is fine-tuning pretrained encoder-decoder models, for example, BART~\cite{lewis-etal-2020-bart} and T5~\cite{JMLR:v21:20-074-t5}.
Recently, interest has shifted to zero-shot or few-shot generation using decoder-only models such as GPTs~\cite{NEURIPS2020_1457c0d6-gpt}.
Several prompting methods have been proposed for the table-to-text and graph-to-text settings~\cite{axelsson-skantze-2023-using,lorandi-belz-2023-data}; however, none have been proposed for the time-series numerical-data-to-text setting.
% linearization手法
% ベストプラクティスが明らかではないことを言う。
% タスク指向対話での発話生成。構造化データを自然言語ぽくするアイデア。https://aclanthology.org/2020.emnlp-main.527.pdf
% droneのセンサーデータ: https://aclanthology.org/2022.lrec-1.745.pdf

% 時系列数値データをlinearizeする場合の問題・課題・難しさ
% 時系列数値データをlinearizeする研究としてはdroneデータの研究がある (https://aclanthology.org/2022.lrec-1.745.pdf)
Time-series numerical data can be converted into tables by aligning numerical values and timestamps.
In this case, the linearization method proposed in the existing table-to-text research can be applicable.
In addition, expressions in programming languages such as Python have been used in research on automatic code generation.
In this study, we propose using time-series numerical sequences as such representations.

\section{Task}
This section describes the task settings.
Figure~\ref{fig:input_format} presents an example of an input and its corresponding output market comments.
We use two time-series numerical sequences derived from the Nikkei Stock Average (Nikkei225): 1) a long-term series that records daily closing prices over the last seven days, and 2) a short-term series that captures daily price fluctuations at five min intervals from market opening to closing.
The system generates a market comment on the target timestamp based on these two inputs.

\section{Methods}
\label{sec:methods}
This section compares the prompting methods.
We assume that each number in sequences can be aligned to a timestamp.
In the example shown in Table \ref{tab:converted_sequence}, a value 9988.05 yen in an input time-series numerical sequence can be aligned to, for example, the timestamp of 15:00, at which the value was tracked.

Table \ref{tab:prompt-example} lists the prompt templates we use for zero- and few-shot generations.
In practice, we replace \texttt{[INPUT FORMAT(short-term)]} and \texttt{[INPUT FORMAT(long-term)]} with the respective input formats in Tables~\ref{tab:example-input-format-short}~and~\ref{tab:example-input-format-long}.
This prompt contains two placeholders, i.e., \texttt{[INPUT FORMAT] (short-term)} and \texttt{[INPUT FORMAT] (long-term)}, which are replaced by one of the nine prompts listed in Tables~\ref{tab:example-input-format-short}~and~\ref{tab:example-input-format-long}.
These prompts are divided into four categories: 1) direct prompts; 2) converting the sequence into a table and then linearizing it; 3) converting the sequence into expressions used for computer programs; and 4) filling a template that produces human-like language.

\subsection{Direct Prompt (Baseline)}

This method simply combines adjacent numerical values with a space to represent a time-series numerical sequence as a sequence of space-separated numerical values.

\subsection{Linearized Table}

\begin{table}[t]
    \centering
    \small
    \begin{tabular}{l|l}
    \toprule
    Timestamp & Nikkei225 \\ \hline\hline
    15:00 & 9988.05 \\\hline
    14:55 & 9982.06 \\\hline
    14:50&  9978.11 \\\hline
    14:45 & 9972.66 \\\hline
    14:40&  9967.11\\\hline
    14:35 & 9961.37\\ 
    \bottomrule
    \end{tabular}
    \caption{Example of a table converted from a numerical sequence of market prices.}
    \label{tab:converted_sequence}
\end{table}

A direct prompt does not use the information about the alignment between a value and its timestamp.
Thus, we promptly included richer information by adding more information about the alignments.
We convert a time-series numerical sequence into a table as shown in Table~\ref{tab:converted_sequence}.
Each value in the input sequence is aligned with the timestamp at which the market price is recorded.
The converted table has two rows: the first row contains a sequence of timestamps, and the second row represents a sequence of market prices.
Each row has the headings ``Time'' and the name of the market index, i.e., ``Nikkei225'', respectively.

Next, we apply linearization inspired by existing table-to-text studies.
In this category, we compare two linearization methods: \texttt{Column} and \texttt{Row} as explained below:

\begin{description}
   \item[Column] This method extracts the heading and values for each column and concatenates them by adding spaces. This process is performed for two columns, and the two token sequences are combined by placing a space to obtain the final column.
\item[Row] As with the \texttt{Column} method, a sequence of space-delimited tokens is obtained for each column, and the two series are joined by a breakline symbol (``{\it  \textbackslash n}''). This includes the information that the two token sequences are generated from different columns.
\end{description}

\begin{table*}[t]
    \centering
    \scriptsize
    \begin{tabular}{l|rrr|rrr|rrr}
    \toprule
    & \multicolumn{3}{c|}{\bf{0-shot}} & \multicolumn{3}{c|}{\bf{5-shot}} & \multicolumn{3}{c}{\bf{10-shot}} \\
        & \bf{BLEU} & \bf{METEOR} & \bf{BERTScore} & \bf{BLEU} & \bf{METEOR} & \bf{BERTScore} & \bf{BLEU} & \bf{METEOR} & \bf{BERTScore} \\
    \midrule 
    Direct &   0.01 & 0.48 & 60.30 & 8.26 & 25.22 & 73.50 & 9.39 & 26.55 & 73.96   \\
    \midrule 
    Column &   0.38 & 14.06 & 65.33 & 8.30 & 24.99 & 73.35 & 9.49 & 26.00 & 73.65   \\
    Row &   0.42 & 8.86 & 64.83 & 9.16 & 26.33 & 73.76 & 10.49 & 27.88 & 74.31   \\
    \midrule 
    Python List &   0.36 & 16.16 & 65.01 & 8.32 & 25.32 & 73.54 & 9.59 & 26.51 & 73.87   \\
    Python List~(nested) &   0.40 & 8.94 & 65.77 & 9.15 & 26.77 & 74.01 & 9.86 & 27.42 & 74.15   \\
    Python Dictionary &   0.44 & 9.60 & 65.40 & 9.17 & 26.42 & 73.96 & 10.41 & 28.25 & 74.56   \\
    HTML Table &   0.35 & 12.26 & 63.93 & 8.30 & 26.10 & 73.92 & 8.45 & 26.44 & 74.08   \\
    LaTeX Table &   0.44 & 15.11 & 67.59 & 8.36 & 26.10 & 73.76 & 9.53 & 27.56 & 74.02   \\
    \midrule 
    Text~(English) &   0.30 & 15.66 & 60.13 & 8.49 & 25.92 & 73.97 & 9.10 & 26.95 & 74.52   \\
    Text~(Japanese &   0.03 & 0.95 & 55.35 & 8.51 & 26.55 & 74.00 & 9.26 & 27.60 & 74.21   \\
    \bottomrule  
    \end{tabular}
    \caption{Comparison of methods in terms of BLEU, METEOR, and BERTScore.}
    \label{tab:my_label}
\end{table*}

\begin{table}[t]
    \centering
    \scriptsize
    \begin{tabular}{l|rrr}
    \toprule
        & \bf{BLEU} & \bf{METEOR} & \bf{BERTScore}  \\
    \midrule 
    EncDec & 11.41& 30.90 &75.94 \\
    \midrule
    Direct &   9.39 & 26.55 & 73.96   \\
    \midrule 
    Column &   9.49 & 26.00 & 73.65   \\
    Row &   10.49 & 27.88 & 74.31   \\
    \midrule 
    Python List &  9.59 & 26.51 & 73.87   \\
    Python List~(nested)  & 9.86 & 27.42 & 74.15   \\
    Python Dictionary &    10.41 & 28.25 & 74.56   \\
    HTML Table &   8.45 & 26.44 & 74.08   \\
    LaTeX Table &    9.53 & 27.56 & 74.02   \\
    \midrule 
    Text~(English) & 9.10 & 26.95 & 74.52   \\
    Text~(Japanese &  9.26 & 27.60 & 74.21   \\
    \bottomrule  
    \end{tabular}
    \caption{Comparison of prompts using 10-shot and EncDec in terms of BLEU, METEOR, and BERTScore.}
    \label{tab:my_label}
\end{table}

\subsection{Programming Language-like Prompts}

We compare the idea of adopting the representation methods used in programming languages, such as Python List and Dictionary.
LLMs are also pre-trained on source code extracted from repositories such as GitHub.
The inclusion of source-code in a prompt is common, especially in source code generation tasks, in which high performance in both understanding and generation has been reported.
Therefore, it is beneficial to convert an input time-series numerical sequence into a form similar to that of a programming language.

\begin{description}
   \item[Python List] This method first creates two python-like lists named ``{\it Time}'' and ``{\it Nikkei225}''.
   The former list contains the timestamps as string values, e.g., {\it ["15:00", "14:55", "14:40", ... "14:35"]}.
   The latter list contains the numerical stock prices as floating values {\it [9988.05, 9982.06, ..., 9961.37]}.
   Finally, these two lists are concatenated by a space.
   \item[Python List (nested)] This prompting method first creates a python-like list with two elements for each timestamp and price pair, e.g., {\it [15:00, 9988.05]} means that the price is 9988.05 at 15:00.
  This method iteratively adds the created list into another python list {\it Nikkei225}, as listed in Table 1.
   \item[Python Dictionary] This method converts a numerical sequence into a Python Dictionary format with key-value pairs, where each key represents the timestamp and each value corresponds to a stock price e.g., \{"15:00": 9988.05, "14:55": 9982.06, ...\}.
   \item[HTML] As with the \texttt{ Column} method, \texttt{ HTML} method assumes that the time-series numerical sequence is represented by a two-row table. This method represents a table as an HTML code, as shown in Table 1.
   \item[LaTeX] As another format for representing a table, we also compare the Latex format.
\end{description}

\subsection{Language Template-based Prompt}

LLMs are trained mainly on natural language text on the Web.
We hypothesize that prompts resembling natural language work well.
Thus, we propose a template-based method for converting a time-series numerical sequence into a natural language-like-prompt.
Specifically, we use two templates: \texttt{ Text (English)} and \texttt{ Text (Japanese)}, as explained below:

\begin{description}
   \item[Text (Japanese)]  This setup uses the template {\it ``AAA時点での日経平均はBBB。''}, which means {\it ``Nikkei225 as of AAA is BBB yen''}. 
   In this template, AAA refers to the timestamp and BBB is the corresponding market price. 
   For each time-price pair, we obtain a sentence by filling the slots. 
   All obtained sentences are finally combined by a breakline symbol (``{\it  \textbackslash n}'').
   \item[Text (English)] For many released LLMs, a large portion of the training data for pretraining is in English. Thus, we also compare the English version of the prompt, i.e., {\it ``Nikkei225 as of AAA is BBB yen''}.
\end{description}

\section{Experiments}

Here, we describe our experiments; dataset, comparison methods, and evaluations.

\subsection{Dataset}
%\cite{murakami-etal-2017-learning}に従い，日経平均株価データと日本語で書かれた市場コメントデータが対になったデータセットを作成した． 例をFigure?に示す． このデータはIBI-Square Stocksから取得し，December 2010からseptember 2016の日経平均株価のデータが含まれている．
%合計で18,489データを取得し，train/valid/testデータとして，それぞれ15,035/1,759/1,695データに分割した． 入力データとして，5分おきで取得されたshortデータと前日までの７日間の株価の終値であるlongデータを入力データとして用いた． 取引市場は9:00から11:30の午前の部と12:30から15:00の午後の部に分かれているため，short dataの長さは62であり，long dataの長さは7である． 
We use the dataset in \citet{murakami-etal-2017-learning, aoki-etal-2018-generating, hamazono-etal-2021-unpredictable}. 
The dataset contains 18,489 pairs of time-series numerical sequences and market comments. 
The time-series numerical sequences are obtained from IBISquare\footnote{\url{http://www.ibi-square.jp/index.html}}. 
Stock price data are collected from December 2010 to September 2016. 
The comments are provided by Nikkei Quick News\footnote{"We have released the source code for preprocessing at \url{https://github.com/aistairc/market-reporter}.}.
We split this dataset into training, validation, and testing sets, comprising 15,035, 1,759, and 1,695 data points, respectively. For this study, we define ``short-term sequence'' as the sequence of stockvalues recorded every five minutes and ``long-term sequence'' as the closing prices for the stock over the last seven days. The Nikkei market operates in two sessions: 1) a morning session from 9:00 to 11:30, and 2) an afternoon session from 12:30 to 15:00. Consequently, the short-term sequence comprised a maximum of 62 prices per day, whereas the long-term sequence comprises seven days.

\subsection{Compared Methods}
%baselineモデルとして，BARTベースのfine-tuningモデル用いる．
%先行研究の\cite{murakani,aoki,hamazono}の方法に従い，

We compare our zero- and few-shot models with an existing fine-tuned encoder-decoder model~\cite{murakami-etal-2017-learning}.

\subsubsection*{Existing Encoder-decoder Model~(EncDec)}
We use an extended implementation of an existing model by~\citet{murakami-etal-2017-learning}.
The original implementation directly encodes long- and short-term time-series numerical sequences using multilayer perceptrons (MLPs), and then, the encoded vectors are fed into an LSTM-based encoder-decoder.
Our extended implementation uses a pretrained BART~\footnote{\url{https://huggingface.co/stockmark/bart-base-japanese-news}} for the encoder-decoder, which has been used more frequently in recent data-to-text settings.
Our implementation uses multilayer perceptrons~(MLPs) to convert both short-term and long-term input vectors into fixed-size vectors of size 768.
This size is selected because the embedding layer of the BART is 768.
Finally, BART then receives the two vectors and outputs the comments.
This model is fine-tuned on the abovementioned dataset to minimize the cross-entropy loss.

\begin{figure*}
    \centering
    \includegraphics[width=\linewidth]{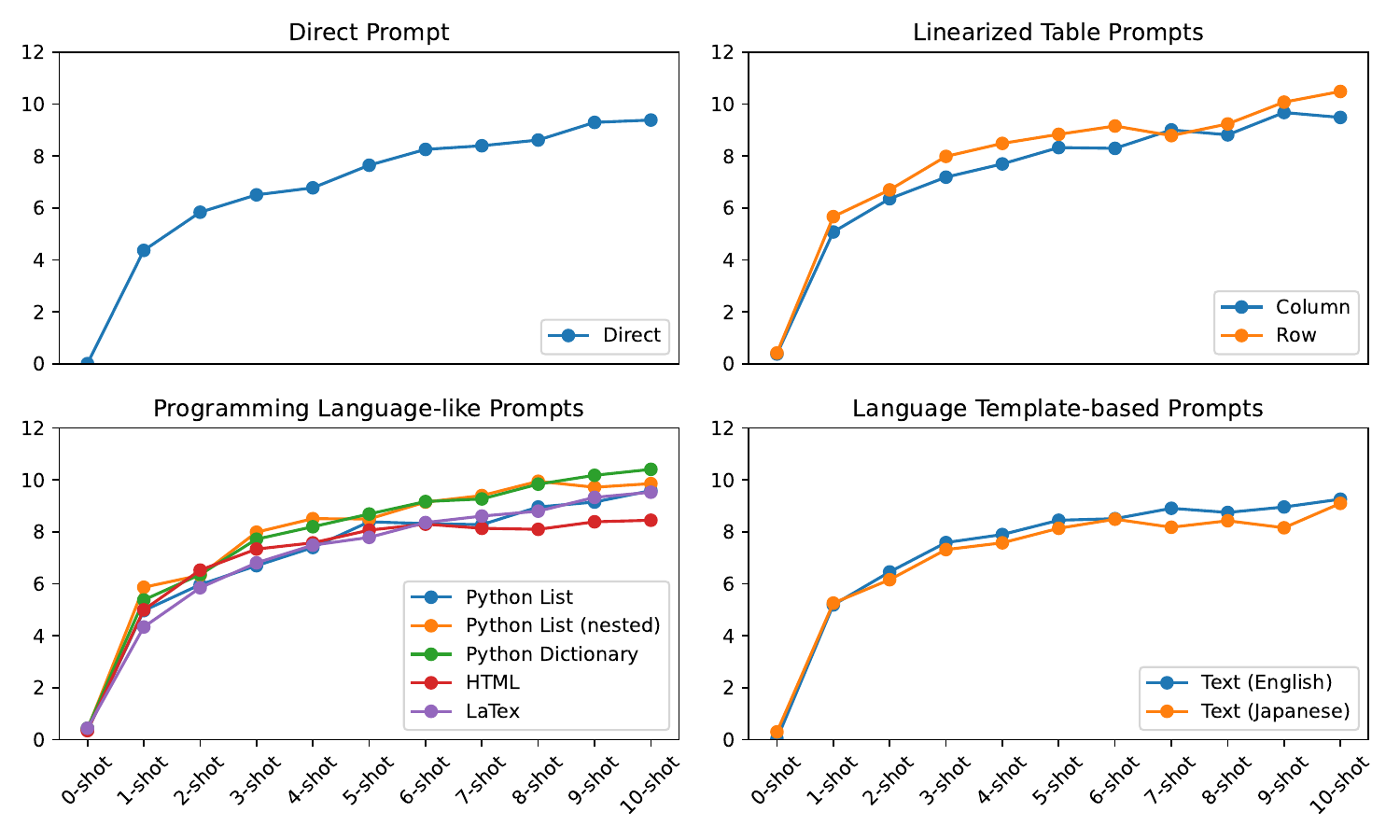}
    \caption{BLEU scores of different prompts on different numbers of shots.}
    \label{fig:all_shot_bleu}
\end{figure*}

\begin{figure}
    \centering
    \includegraphics[width=\linewidth]{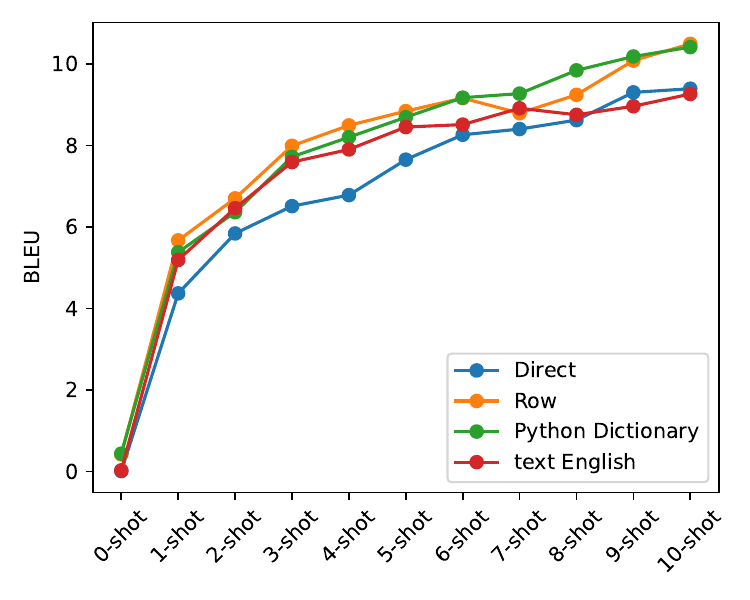}
    \caption{BLEU scores of different prompts on different numbers of shots.}
    \label{fig:best_each_method}
\end{figure}

\subsubsection*{Proposed Zero- and Few-shot Models}

We use an instruct-tuned GPT~\cite{NEURIPS2020_1457c0d6-gpt}\footnote{In particular, we use {\it gpt-3.5-turbo} in OpenAI's API.}.
% few-shot事例の作り方
In the zero-shot setting, the prompt shown in Table~\ref{tab:prompt-example} is used as is.
In the few-shot setting, we append a reference comment to the prompt after {\it Output:} in the prompt.
We compare up to 10 shots in our experiments.
We set the maximum number to 10 owing to the length limits of the GPT.
In preliminary experiments, we found that the performance varied significantly depending on the instances used for the shots.
Therefore, we randomly select the instances for shots, repeat the experiments 10 times, and report the averaged scores.

\subsection{Automatic and Human Evaluations}
In this subsection, we describe the evaluation.

\subsubsection{Metrics for Automatic Evaluation}

For the scores, we adopt the commonly used BLEU and F1-score from the BERTScore.
BLEU has been employed in many existing studies; however, further evaluation is required because it considers only surface words.
Therefore, BERTScore, which uses neural network embedding to capture semantic similarities, is employed.

Such automatic evaluation metrics only capture the similarity between automatically generated and reference comments but do not capture the correctness of the generated text.

\subsubsection{Evaluation by Human Judges}

Therefore, we also evaluate the generated comments by human judges.
For each comparison method, we present 30 comments of time-series numerical data to a human evaluator, who is a native Japanese speaker, and ask the evaluator to evaluate whether the output text is consistent or inconsistent with the reference.
The evaluator can also refer to the short-term and long-term sequences, if needed.
We report the number of consistent and inconsistent results for each method.

\section{Results and Discussions}

The results are presented in the following section.

\subsection{Automatic Evaluation}

Figure \ref{fig:all_shot_bleu} shows the changes in BLEU scores with the number of shots.
The detailed scores of BLEU, METEOR, and BERTScore are shown in Table \ref{tab:my_label}.
Overall, the scores generally increase with the number of shots, except for the \texttt{HTML} prompt, as shown in the bottom-left graph in Figure \ref{fig:all_shot_bleu}.
All methods on the zero-shot setting achieve very low scores in terms of BLEU, METEOR, and BERTScore, implying that the multitasking capability of the LLM is not sufficient for the market comment generation.
\texttt{Direct} prompt does not work well for this task, and all other prompts outperform it.

A further look at the scores for prompts other than the \texttt{Direct} prompt yielded three findings: 1) \texttt{Python Dictionary}, \texttt{Python List (nested)}, and \texttt{Row} work better than the other prompts, 2) \texttt{Text} unexpectedly works worse; and 3) among programming language-based prompts, \texttt{HTML} performs the worst.
Each finding is explained in detail below.

\subsection*{Finding 1: \texttt{Python Dictionary}, \texttt{Python List (nested)}, and \texttt{Row} perform better}

Table \ref{tab:my_label} lists the BLEU, METEOR, and BERTScores for each prompt in all categories.
Owing to the space limitation, we show the values obtained in the setting of 0, 5, and 10 shots.
Overall, we observe that there are three best-performing models among all models: \texttt{Python Dictionary}, \texttt{Python List (nested)}, and \texttt{Row}.
These best-performing prompts share common characteristics: 1) they have similar characteristics to the data used for pretraining, and 2) the timestamps and market prices are written closely to each other in a prompt.
Python codes are also used for the pretraining of the LLMs, thus, better results may be reasonable.
\texttt{ Row} also achieves better results because this approach is used for dataset creation for pretraining to linearize HTML tables on the web.

Furthermore, the timestamps and market prices are written closely to each other, for example, {\it [15:00, 9988.05]} in the \texttt{ Python List (nested)} and {\it 15:00":9988.05} in the \texttt{ Python Dictionary}.
Other programming language-like prompts, such as \texttt{ Python List}, \texttt{ HTML}, and \texttt{ LaTeX}, with lower performance write the timestamps and the market prices far away from each other.
One possible reason for this is that GPT was able to accurately capture the correspondence between the prices and a time.

\subsection*{Finding 2: \texttt{ Text} unexpectedly performs worse}

We hypothesized that \texttt{ Text (English)} and \texttt{ Text (Japanese)} would perform better; however, Table \ref{fig:best_each_method} reveals that these methods perform worse than the best-performing programming language-like prompts, that is, \texttt{ Python Dictionary} and \texttt{ Python List (nested)}.
Specifically, the performances of \texttt{Text (English)}, represented by a red line, are lower than those of other prompts, that is, \texttt{Row} and \texttt {Python Dictionary}, which are defined by orange and green lines, respectively, except for 7-shot setting for the \texttt{Row} prompt.

One possible reason for this lower performance is that the template generates repetition of the same expression.
For example, the English template generates ``Nikkei225 as of 15:00 is 9982.06 yen'' and repeatedly generates similar expressions.
Such text is unnatural and may not be included in the web texts used for pretraining.

Interestingly, the bottom left graph in Table~\ref{fig:all_shot_bleu} shows that \texttt{Text~(English)} performs better than \texttt{Text~(Japanese)} although the task is to generate a comment in Japanese.
A possible reason could be that English texts are used more frequently during pretraining.

\begin{table*}
    \CJKfamily{ipxma}
    \centering
    \footnotesize
    \begin{tabularx}{\textwidth}{lX}
    \toprule
    Reference &  日経平均、\textcolor{red}{続伸} 前引けは421円\textcolor{red}{高}の１万6349円\\
    & \textit{(The Nikkei 225 \textcolor{red}{extended its gains}, closing \textcolor{red}{up} 421 yen at 16,349 at the midday break)}\\
    \midrule
    \midrule
    \addlinespace[0.1cm]
    EncDec & 東証前引け、\textcolor{red}{反落} 米株安や円安一服で 利益確定売りも \\
    & \textit{(Tokyo stocks \textcolor{red}{fall} at midday close, retreating on Wall Street declines and a pause in yen weakening, with profit-taking also in play.)}\\
    \addlinespace[0.1cm]
    Direct &  日経平均、\textcolor{red}{反落} 前引けは80円\textcolor{red}{安}の１万6349円\\
    & \textit{(The Nikkei 225 \textcolor{red}{reversed course}, \textcolor{red}{falling} 80 yen to close at 16,349 at the midday break.)}\\
    \midrule
    \addlinespace[0.1cm]
    Column &  日経平均、\textcolor{red}{反落} 前引けは421円\textcolor{red}{安}の１万6349円\\
    & \textit{(The Nikkei 225 \textcolor{red}{reversed course}, \textcolor{red}{falling} 421 yen to close at 16,349 at the midday break.)}\\
    Row&  日経平均、\textcolor{red}{反発} 前引けは421円\textcolor{red}{高}の１万6349円 \\
    & \textit{(The Nikkei 225 \textcolor{red}{rebounded}, closing \textcolor{red}{up} 421 yen at 16,349 at the midday break.)} \\
    \midrule
    \addlinespace[0.1cm]
    Python List & 日経平均、\textcolor{red}{反落} 前引けは421円\textcolor{red}{安}の１万6349円  \\
    & \textit{(The Nikkei 225 \textcolor{red}{reversed course}, \textcolor{red}{falling} 421 yen to close at 16,349 at the midday break.)}\\
    \addlinespace[0.1cm]
    Python List~(nested)& 日経平均、\textcolor{red}{大幅続落} 前引けは421円\textcolor{red}{安}の１万6349円 \\
    & \textit{(The Nikkei 225 \textcolor{red}{continued to drop sharply}, \textcolor{red}{falling} 421 yen to close at 16,349 at the midday break.)}\\
    \addlinespace[0.1cm]
    Python Dictionary  & 日経平均、\textcolor{red}{反落} 前引けは421円\textcolor{red}{安}の１万6349円 \\
    & \textit{(The Nikkei 225 \textcolor{red}{reversed course}, \textcolor{red}{falling} 421 yen to close at 16,349 at the midday break.)}\\
    \addlinespace[0.1cm]
    HTML &  日経平均、\textcolor{red}{続伸} 前引けは421円\textcolor{red}{高}の１万6349円 \\  
    & \textit{(The Nikkei 225 \textcolor{red}{extended its gains}, closing \textcolor{red}{up} 421 yen at 16,349 at the midday break.)}\\
    LaTeX &  日経平均、\textcolor{red}{続伸} 前引けは200円\textcolor{red}{超}の１万6349円\\
    & \textit{(The Nikkei 225 \textcolor{red}{extended its gains}, closing \textcolor{red}{up} 200 yen at 16,349 at the midday break.)}\\
    \midrule
    \addlinespace[0.1cm]
    Text~(English)  &   日経平均、\textcolor{red}{反落} 前引けは121円\textcolor{red}{安}の１万6349円  \\
    & \textit{(The Nikkei 225 \textcolor{red}{reversed course}, \textcolor{red}{falling} 121 yen to close at 16,349 at the midday break.)}\\
    \addlinespace[0.1cm]
    Text~(Japanese)  &    日経平均、\textcolor{red}{反落} 前引けは421円\textcolor{red}{安}の１万6349円  \\
    & \textit{(The Nikkei 225 \textcolor{red}{reversed course}, \textcolor{red}{falling} 421 yen to close at 16,349 at the midday break.)}\\
    \bottomrule
    \end{tabularx}
    \caption{Comparison of text output by each method, with 'movement terms' highlighted in red in the table.}
    \label{tab:outout_text}
\end{table*}

\subsection*{Finding 3: Among programming language-based prompts, \texttt{ HTML}  and \texttt{ LaTeX} perform worse}

The bottom left graph in Figure \ref{fig:all_shot_bleu} shows each model's BLEU scores in the program language-like prompts category.
Among the prompts in this category, \texttt{ HTML}  and \texttt{ LaTeX} perform worse than other prompts.
In particular, the performance of \texttt{ HTML} has not improved after 5-shot and has already converged, whereas the performances of the other prompts are still improving.

According to the statistics provided by GitHub\footnote{\url{https://madnight.github.io/githut}}, one of the pretraining sources, the proportion of HTML formatted files is less than 0.1\%.
One possible reason may be the small percentage of data used for the pretraining.
Furthermore, the \texttt{HTML} prompt is inherently longer than the others, which makes reasoning using by the GPT difficult.

\begin{table}[t]
    \centering
    \scriptsize
    \begin{tabular}{l|rrr}
    \toprule
        & \bf{\# Consistent} & \bf{\# Inconsistent} &  \\
    \midrule 
    EncDec & 7 & 23 \\ 
    Direct &   8 & 22 \\
    Row &  \bf{14}  & 16  \\
    Python Dictionary &  12   & 18   \\
    Text~(English) & 12 &   18  \\
    \bottomrule  
    \end{tabular}
    \caption{Results of human evaluation on EncDec, Direct, Row, Python Dictionary, and Text English.}
    \label{tab:human_eval}
\end{table}

\subsection{Evaluation by Human Judges}

We conducted human evaluations of the encoder-decoder method and the method with the highest BLEU score for each category: \texttt{Direct}, \texttt{Row}, \texttt{Python Dictionary}, and \texttt{Text~(English)}.

Table~\ref{tab:human_eval} presents the results of the human evaluation.
Across all methods, we discovered that the evaluator judges the generated comments more frequently than comments consistent with the reference comments. 
This includes many cases in which the reference mentions that the stock price is rising, but the generated comment states that it is falling, or vice versa.
Table~\ref{tab:outout_text} shows the output of each method.
In this table, the `movement term', referring to the stock price movement such as `rising' or `falling', is highlighted in red.

Interestingly, although the fine-tuned encoder-decoder method achieved high scores in the automatic evaluation, it received lower ratings in the human evaluation.
By contrast, the prompt-based method yielded comments that were more consistent with the reference comments.
We hypothesize that this discrepancy arises because the fine-tuned comments are stylistically similar to the references but do not accurately capture numerical movements. In contrast, the prompt-based method excels in understanding numerical trends but is limited in covering all text styles, given its reliance on at most ten examples as shots. 
To verify this hypothesis, a separate evaluation focusing on the accuracy of references to text style and numerical movements is necessary, which indicates a potential direction for future research.

\section{Conclusion}

In this study, we address the use of LLMs for text generation from numerical sequences through the generation of market comments.
We approached the problem by converting numerical sequences into ten different formats across four categories, and conducted experiments to compare these methods with the existing encoder-decoder approach based on BART. 
Our results indicated that prompts based on programming languages yielded strong performance, whereas textual prompts and those based on HTML and LaTeX tables, were less effective.
Furthermore, a human evaluation assessing the consistency with reference revealed that the prompt-based methods outperformed the encoder-decoder method. 
This highlights the potential of prompt-based approaches to generate market comment and similar data-to-text tasks.

Although our research focused on market comments as a representative task with numerical sequence inputs, many other tasks share comparable characteristics.
In future work, it will be valuable to conduct experiments on datasets with other types of numerical sequence inputs to further validate and extend our findings.

\section*{Acknowledgements}
This study is based on the results obtained from a project JPNP20006, commissioned by the New Energy and Industrial Technology Development Organization (NEDO). 
Ideas related to time-aware prompts are obtained from discussions in the BRIDGE (Program for Bridging the Gap Between R\&D and Ideal Society (Society5.0) and Generating Economic and Social Value) supported by Cabinet Office.
For experiments, the computational resource of the AI Bridging Cloud Infrastructure (ABCI) provided by the National Institute of Advanced Industrial Science and Technology (AIST) were used.

%\section{Language Resource References}

\bibliographystylelanguageresource{lrec-coling2024-natbib}
\bibliographylanguageresource{languageresource}

%\nocite{*}
\section{Bibliographical References}\label{sec:reference}

\bibliography{paper}
\bibliographystyle{lrec-coling2024-natbib}

\end{document}